# Can LLMs Help Improve Analogical Reasoning For Strategic Decisions? Experimental Evidence from Humans and GPT-4

**Phanish Puranam**

Professor of Strategy, INSEAD Singapore

phanish.puranam@insead.edu

**Prothit Sen**

Assistant Professor of Strategy, Indian School of Business, Hyderabad

Prothit_Sen@isb.edu

**Maciej Workiewicz**

Associate Professor of Management, ESSEC Business School

workiewicz@essec.edu

**Abstract**

This study investigates whether large language models (LLMs), specifically GPT-4, can match human capabilities in analogical reasoning for strategic decision-making. Using a novel experimental design that requires source–target matching, we find that GPT-4 achieves high recall—retrieving all plausible analogies—but suffers from low precision, frequently applying incorrect analogies based on superficial surface features. Humans, by contrast, exhibit high precision but low recall, selecting fewer analogies yet with stronger causal alignment. These findings advance theory by highlighting *matching*—the evaluative phase of analogical reasoning—as a distinct and critical step that requires accurate causal mapping over and above mere retrieval of an analogue. While current LLMs excel at retrieving analogies, human cognitive superiority persists in mapping causal structures across domains. Error patterns further reveal that AI failures stem from superficial similarity detection, whereas human errors reflect subtler misinterpretations of causal logic. In aggregate, the findings suggest a potentially productive division of labor in organizations: AI can serve as an analogy generator, while human decision makers act as critical evaluators in the application of the most contextually relevant causal schemas to organizational problem solving.

# 1. INTRODUCTION

Analogical reasoning—where managers draw on solutions from past experiences when dealing with novel challenges—is one of the central pillars of strategic thinking (Gavetti, Levinthal, & Rivkin, 2005; Miller & Lin, 2015). Analogies serve as cognitive tools that enable the interpretation of complex or ambiguous situations by linking them to familiar precedents (Gentner, Holyoak, & Kokinov, 2001). They are helpful in a wide range of strategic choice situations, including decisions about market entry, acquisitions, business model innovation, and organizational turnaround (Gavetti, Levinthal, & Rivkin, 2005). Beyond guiding choice, analogies also structure how managers mentally represent their strategic environment. By highlighting certain features and suppressing others, analogies influence how opportunities and threats are construed, thereby shaping the contours of strategic problem framing (Gary, Wood, & Pillinger, 2012).

When Charlie Munger famously remarked, "You've got to have models in your head… and you've got to array your experience—both vicarious and direct—on this latticework of models",[1] he was arguing that analogies from another domain can act as a form of guidance (a model) in ambiguous and data-poor domains. However, his statement about the virtues of possessing a repository of analogies also highlights an important challenge: one must *select* the appropriate analogy from a broader set of possible candidates—a challenge known as the 'matching problem' (Cummins, 1992; Gentner, Rattermann, & Forbus 1993). This is especially critical in strategy, where decisions are high-stakes and often irreversible (Leiblein, Reuer, & Zenger, 2018). A poorly matched analogy can lead to flawed inference, misplaced confidence, and costly strategic errors.

---

[1] https://ryanstemen.medium.com/charlie-munger-was-right-why-loading-your-head-with-mental-models-is-a-superpower-f5742ecc7a88

Matching correctly—often in a one-shot setting with no opportunity for feedback from implementation—requires not just an ability to recall past experiences, but also sophisticated judgment in selecting the appropriate one. Without accurate matching, a large store of possible analogies may, in fact, be a disadvantage. While stories of successful analogical transfer abound in the business world, it remains unclear whether these are indicative of the general power of analogies or simply the visible survivors, overlooking instances where analogical reasoning may have misfired. For instance, while Kodak executives successfully modeled their film and camera business on the analogy to how Gilette sold razors and razor blades, they also considered digital images to be analogous to traditional film, a mistake they did not recover from (Tripsas & Gavetti, 2000). Wayne Huizenga successfully leveraged his experience in the funeral homes business, to consolidate local video rental businesses into Blockbuster. But eventually Blockbuster paid the price for mistakenly viewing the Netflix business model as an analog to its own DVD distribution model. Overall, how human decision makers identify good matches and filter out bad analogies is an important and yet poorly understood phenomenon (Blanchette & Dunbar, 2001; Gentner & Smith, 2013)

Rapid developments in Large Language Model (LLM) technologies are of relevance to the challenge of effective analogical reasoning. At its core, the transformer architecture (Vaswani et al., 2017) of the deep learning models that underlie LLMs allows them to detect similarity and relevance between ideas expressed as text, i.e., between a given prompt and patterns embedded in their training data. However, similarity detection is necessary, but not sufficient, for effective analogical reasoning (Olguin, Tavernini, Trench, Ricardo, & Minervino, 2022). Even if LLMs can outperform humans in the retrieval of potential analogies based on similarity, their ability to map structural similarities effectively to produce good matches remains unproven. On the dimension of structural abstraction, particularly in the domain of verbally formulated reasoning, as is common in business contexts, there is scant evidence on the relative performance of LLMs and humans (Yuan et al., 2023; for non-verbal reasoning see Lewis & Mitchell, 2025; Camposampiero et al., 2025). In other words, are LLMs capable of performing analogy retrieval and matching, and how does their performance compare with that of humans? Can human decision makers benefit from analogical reasoning executed by LLMs? What are the complementarities between analogical reasoning by human and AI agents?

This paper directly investigates these questions by comparing the verbal analogical reasoning abilities of GPT-4 and humans in a business context. In a novel experimental setup designed to replicate and meaningfully extend—classic work on analogical transfer, we introduced a matching problem by pairing two source analogs with two target problems that reflect typical business contexts. Human participants (n = 199) and GPT-4 (n = 60 independent trials) were tasked with solving each target problem under both 'no hint' and 'hint' conditions, where a hint made the existence of a possible analogical frame explicit, without pointing out the correct one. Accuracy was measured by whether responses solved the problem, as pre-specified using causal schema diagrams, and matching was measured by eliciting which source story had been used to solve it.

Our findings reveal a nuanced picture. While GPT-4 outperforms human participants in correctly solving the problem, its use of analogical reasoning is less effective than that of humans.. A confusion matrix analysis reveals a critical trade-off: while GPT-4 has high recall (it almost never misses a relevant analogy), it suffers from poor precision (frequent false positives). Humans, conversely, show high precision (they seldom misapply an analogy), but poor recall (they often fail to recognize the correct one). In other words, GPT-4 behaves like an eager classifier—sensitive but noisy—while humans act as cautious diagnosticians—accurate but reserved. As a result, GPT-4 achieves lower overall accuracy on the matching task despite appearing to apply analogical transfer more often. Moreover, our experiments reveal that GPT-4's analogical errors stem largely from surface-level similarity detection, while human errors more often reflect incorrect abstraction of structural schemas. This highlights not just a performance gap, but a fundamental difference in the failure modes of analogical reasoning between humans and LLMs.

Our findings imply that the distinctive pattern of performance of humans and AI in analogical reasoning—with the former performing better at mapping and the latter producing more candidates through retrieval—can be harnessed through a collaborative structure; LLMs may dominate humans in generating potential candidate analogies, humans dominate the LLMs in screening for structural relevance. While the next generation of LLMs will doubtless improve on the current ones, the current superiority of LLMs at generating a larger set of candidate analogies will most likely be preserved. It remains an open question whether their capacities at detecting structural similarity will eventually exceed that of humans (Yuan et al., 2023; Olguin et al., 2022); but till it does, human (managerial) involvement in analogical reasoning will be necessary for successful analogical transfer to aid strategic decision-making in organizations.

## 2. BACKGROUND LITERATURE

Analogical reasoning plays a vital role in strategic and organizational decision-making, particularly under conditions of novelty, ambiguity, and incomplete information (Miller & Lin, 2014). Managers frequently draw analogies from past experiences to frame new challenges (Gavetti, Levinthal, & Rivkin, 2005; Tripsas & Gavetti, 2000). Analogies also shape how managers perceive opportunities and threats (Gary, Wood, & Pillinger, 2012). In entrepreneurial settings, analogies serve to frame new ventures, drawing on familiar categories that guide investor expectations and strategic alignment (Navis & Glynn, 2011; Santos & Eisenhardt, 2009). Similarly, analogical reasoning facilitates capability reconfiguration in dynamic markets by transferring lessons from prior contexts (Helfat & Peteraf, 2003; Zollo & Winter, 2002). Strategy narratives often invoke analogies from war, sports, or ecosystems to make complexity more tractable (Cornelissen, Holt, & Zundel, 2011).

## 2.1. Retrieval and Mapping: the foundations of analogical reasoning

Research in cognitive psychology recognizes analogical reasoning to be a fundamental cognitive process that enables the transfer of knowledge across domains (Gentner, 1983; Holyoak & Thagard, 1989). Researchers have converged on two sub-processes to describe analogical reasoning: retrieval, where possible source analogies come spontaneously to mind, and mapping, where the candidate analogies are compared to the target problem at hand. Underlying both sub-processes are the concepts of mental representations and similarity between them. One can represent two problem domains using networks which capture the key concepts and the causal interconnections between them. The two networks may resemble each other in terms of the nodes used (superficial or surface similarity) or the pattern of relationships between the nodes (structural or deep similarity). Successful analogical reasoning depends on finding structural similarities between situations, such that knowledge of a solution in one situation (source) can act as a useful hypothesis about the second (target) (Gentner & Smith, 2013; Goldwater & Gentner, 2015).

For human decision makers, spontaneous analogical retrieval is often cue-dependent and superficial (Gick & Holyoak, 1980). Similarity detection can occur purely based on superficial features across domains. Each feature in the source domain "activate[s] memory representations of other situations that share that feature." (Holyoak & Koh, 1987; p. 333). Structural mapping involves checking for structural similarity across source and target domains (Gentner, 1983) via the detection of correspondence between "causal relations in the two situations." (Holyoak & Koh, 1987; p. 334). Constraint satisfaction—balancing semantic, pragmatic, and structural coherence—supports accurate mapping (Holyoak & Thagard, 1989).

To illustrate these four concepts- retrieval, mapping, surface, and structural similarity- consider the following hypothetical example: a mid-sized drone manufacturer that is currently evaluating a strategic shift from outright product sales to a Drone-as-a-Service model. Under this proposed model, clients would be charged on a per-flight-hour basis, while the drone manufacturer would retain ownership of the assets and assume responsibility for maintenance and guaranteed uptime. For many managers, the most readily accessible analogy – the one that is *retrieved*- may be the widely recognized razor-and-blade (or printer-and-ink) pricing model. Both cases ostensibly involve durable goods complemented by a stream of recurring revenues, rendering this example cognitively salient due to *surface similarity*. However, a more careful examination of the relational structure reveals critical misalignments: in the printer model, vendors generate margin through the sale of consumables, whereas the drone maker would most likely treat maintenance and spare parts as cost centres while monetizing equipment availability.

In contrast, a less likely to be retrieved but *structurally* more aligned analogue is the "power-by-the-hour" model pioneered by Rolls-Royce in the context of jet engines. In this case, the manufacturer retains ownership, earns revenue in proportion to usage, and bears operational risks traditionally held by the customer, thus representing a deeper match in terms of underlying causal mechanisms. This contrast illustrates that effective analogical reasoning in strategy formation requires more than the retrieval of familiar comparisons and the mapping of surface features; it also demands the *mapping* of candidate analogues based on the alignment of their causal structures with the focal decision context.

In other words, given multiple possible analogs to a problem, retrieval and mapping jointly produce matches, good or bad. Retrieval of the most useful analogy for a situation is not guaranteed, and analogical reasoning can be biased if decision-makers rely on surface-level similarities rather than deeper structural parallels. For instance, successful analogizing between the atom and the solar system depends on identifying that both involve smaller bodies revolving around a central attractor—an abstract causal schema that must be detected and mapped correctly (Gentner & Smith, 2013). This relationship between retrieval and mapping is well captured in the so-called "MAC-Many are called/ FAC-few are chosen" model of analogical reasoning (Forbus, Gentner, & Law, 1995).

Yet, mapping is no guarantor of successful match; good analogies may never be retrieved, and bad ones may not be successfully weeded out. Further, among the retrieved candidate analogies, mapping need not eliminate bad analogies if decision makers satisfice and stick with the first satisfactory analogy rather than iterate through alternatives to assess structural similarity (Gentner & Smith, 2013). Training that emphasizes structural alignment over superficial resemblance improves the quality of analogical reasoning (Gick & Holyoak, 1980). Research also demonstrates that comparative exposure to analogs sharpens structural similarity detection (Vendetti et al., 2015; Lovett & Forbus, 2017; Richland & Begolli, 2016). Prior knowledge and abstraction ability also improve schema recognition and analog selection (Hofstadter, 2001; Goldwater & Gentner, 2015).

Though cognitive scientists are fully aware of the challenges of matching, studies that explicitly set up a matching problem are rare (Gentner & Smith, 2013). Laboratory studies (Gick & Holyoak, 1983; Cummins 1992; Markman & Gentner 1993) typically hand participants a lone source, so they bypass the need to choose among alternatives. Naturalistic observations (Blanchette & Dunbar 2001) record which analogies were selected but cannot measure whether *better* analogies were available or how accurate the choices were. In this study, we design an explicit matching problem in strategic decision making and compare how humans and LLMs fare.

**2.2. Analogical reasoning by machines**

Contemporary approaches to analogical reasoning in machines have evolved significantly, shifting from early rule-based systems toward data-driven methods capable of identifying analogies from large-scale datasets (Pan et al., 2009; Sun & Saenko, 2016). Techniques such as Transfer Component Analysis and Deep CORAL aim to minimize domain divergence by aligning feature distributions, while more recent innovations—including probabilistic analogical mapping—enable forms of zero-shot reasoning by leveraging structural inferences (Webb et al., 2023). In parallel, developments in causal modeling frameworks have foregrounded the importance of relational over surface-level similarity, emphasizing structural isomorphism in analogy construction (Pearl & Bareinboim, 2014; Correa et al., 2022).

However, most current applications of large language models (LLMs) to analogical reasoning have focused on highly constrained tasks. These include pattern completion problems such as non-visual Raven's matrices, digit sequence analogies, molecular property transfer, and multiple-choice science questions—contexts in which a single source-target pair is pre-specified and the model is tasked only with inferring a missing element or selecting among fixed alternatives (Webb et al., 2022; Yuan et al., 2023; Lewis & Mitchell, 2025). While such benchmarks provide useful diagnostics of a model's ability to induce structural patterns (e.g., via structure mapping or abduction), they do not capture the open-ended, linguistically framed analogical tasks that characterize real-world managerial reasoning. Questions such as whether a proposed "Drone-as-a-Service" model more closely resembles a razor-and-blade pricing scheme, or a jet engine leasing arrangement exemplify the type of analogical judgment that existing evaluations bypass.

Equally significant is the neglect of the *matching* problem in current studies. In many settings, candidate analogues are either pre-selected or generated through retrieval mechanisms whose outputs are not subjected to any formal evaluation process. As a result, we lack systematic evidence on whether and how well LLMs can identify the most structurally appropriate source when multiple plausible analogies are available—a decision step that is critical in high-stakes reasoning. Moreover, comparative assessments of human and machine analogical reasoning have concentrated almost exclusively on the mapping stage, leaving both the retrieval and evaluation phases outside the empirical frame.

In this study, we address these three gaps by designing an experimental task that (i) employs naturalistic business vignettes written entirely in free-form language, (ii) embeds an explicit two-analogue matching decision, and (iii) evaluates performance across the full analogical reasoning pipeline—*retrieval*, *mapping*, and *evaluation*—for both humans and LLMs.

## 3. LAB STUDY

**3.1. The experiment:**

The influential studies by Gick and Holyoak (1980; 1983), demonstrated that when participants were given a single source story (e.g., a general attacking a fortress) and asked to solve a target problem (e.g., a doctor treating a tumor), many could apply the source —especially when given a hint. But this one-to-one setup does not involve matching, as it does not test whether people can *discriminate* between multiple, competing potential analogies—some superficially similar, others structurally appropriate. In other words, the Gick and Holyoak studies illuminated how analogical *use* can be elicited (i.e., they are tests of retrieval), but not whether analogical *matching* can be competently performed under realistic cognitive conditions. In our studies, we incorporated a matching problem, where participants were shown two source stories and faced two target problems. Our experiments thus departed from the classical 'radiation problem' studies on two important aspects:

**i) Inducing matching complexity**: In addition to the 'radiation problem' story (Story 1), we introduce an additional story in the source domain (Story 2). Mirroring this in the target domain, we expose the subjects to two problems (Problem 1 and Problem 2) instead of one. To solve Problem 1, Story 1 should serve as the appropriate analogy while Story 2 would act as a placebo. Similarly, to solve Problem 2, Story 2 should serve as the appropriate analogy while Story 1 would act as a placebo. Inducing this complexity to the traditional studies represents the "matching problem" that is at the heart of effective analogical transfer—a process that involves both the retrieval and accurate causal mapping of the right analogue.

**ii) Humans vs. AI**: We run the above set-up for human subjects and independent trials on an established LLM-based AI developed by OpenAI, GPT-4. The human subjects comprise Masters' level students from a leading business school.

### 3.2. Source stories and Target problems:

To introduce matching complexity into the analogical reasoning task—beyond the original Gick & Holyoak (1983) design—we construct an experimental setup with two source stories (S1 and S2) and two target problems (T1 and T2). As experimenters, we know *a priori* that the correct analogical mappings are S1 → T1 and S2 → T2. However, participants are not informed of these mappings, which creates the possibility of incorrect pairings (e.g., S1 → T2 or S2 → T1). This structure allows us to test not just analogy application, but the more cognitively demanding *matching* process that determines analogical transfer.

We retain S1—the radiation problem—from the original Gick & Holyoak studies (1983) to benchmark our results against prior findings. The corresponding target problem, T1, is a novel but structurally analogous scenario in the domain of operations and supply chain management. We design S2 and T2 as a new pair to capture a different type of reasoning challenge—one rooted not in domain-specific knowledge, but in recognizing *survivorship bias* as an abstract causal schema. This bias is broadly applicable across managerial contexts, especially when making inferences from incomplete data (Mundt, Alfarano, & Milakovic, 2022). Based on the above logic, we present two stories as follows:

**Story 1 (the Radiation Problem: Split and converge schema):**

*"Dr. Clarke, a seasoned radiation oncologist, observed with concern as his patient, Sarah, battled severe side effects from her ongoing radiation therapy. Sarah was just one of many patients experiencing the unintended consequences of treating cancerous tumors with high-energy radiation. Although radiation therapy had been a critical component in the fight against cancer for decades, it was not without its flaws. The primary challenge lay in delivering a potent dose of radiation directly to the tumor while sparing the surrounding healthy tissue. Too often, this delicate balance proved difficult to achieve, resulting in collateral damage to nearby organs and exacerbating the patient's suffering. Dr. Clarke knew that a more precise and targeted approach was needed, one that could improve treatment outcomes and enhance the quality of life for patients like Sarah. Dr. Clarke came up with a new technique that involved delivering multiple beams of radiation at varying intensities, aimed at the tumor from different angles. This allowed the cumulative dose of radiation at the tumor site to be sufficient to destroy it, while the surrounding healthy tissue would receive a lower dose, thus minimizing damage."*

**Story 2 (the Dolphins Story: An example of survivorship bias):**

*"Sailors in the 18th and 19th century who were shipwrecked but were lucky enough to be rescued often told stories about how dolphins saved their lives. When they were in the water after their ship capsized, and almost drowning as they became tired, dolphins would gently and playfully nudge them towards an island which had enough fruit and water to keep them alive till they were rescued. But in the 20th century marine biologists who studied dolphins discovered the prosaic (and rather tragic) truth: dolphins are friendly and love playing with people in the water and*

*will nudge them gently- but in all possible directions. Only those sailors who were lucky enough to have been pushed towards an island that allowed them to survive came back to civilization to tell their stories. So, the account of dolphins as the sailor's best friends was based on a biased sample of experience- biased towards survivors."*

**Target domain:** We present two problems such that the solution to Problem 1 should ideally come from the radiation story as analogy and solution to Problem 2 should come from the survivorship bias story of dolphins as an analogy. The following are the two problems:

**Problem 1 (The City Factory problem):**

*"In the thriving coastal city of Industria, a major factory stands as a symbol of economic prosperity. However, a pressing issue looms over this industrial powerhouse. Marine transport became too expensive, and the factory must find a new location to establish a new supply center closer to the city. However, the factory's proximity to the picturesque coastline leaves little room for expansion, and neither the road nor rail infrastructure suffices to meet their entire supply needs. Building an airport is out of the question. Faced with this intricate puzzle, the factory's management must find an innovative solution that will secure a steady volume of supply without overwhelming the city's roadways or railways or compromising the delicate balance between industrial progress and the preservation of the cherished coastal landscape. What could they do?"*

**Problem 2 (The HR problem):**

*"In an effort to enhance productivity, the Human Resources department of a company recently introduced a pilot training program for high-potential employees, focusing on a new agile methodology and its application for marketing. Participation in the program was voluntary, and marketing department employees were allowed to quit at any point if they felt it interfered with their work. As the pilot progressed, participants who completed the entire agile training reported an impressive average productivity increase of 23% and expressed satisfaction with their decision to join the program. Encouraged by these positive results, the company implemented the agile methodology as a mandatory training across the entire organization. The results were disastrous. Employees complained about what a waste of time it was, and average productivity fell by 8%. How could this have happened when the pilot study indicated the training would be good for productivity?"*

It is worth highlighting that inaccurate statistical inferences due to sampling and survivorship bias issues of who gets "selected" and "stays" in the organization is a very common problem plaguing HR policies and practices (Capelli, Tambe, & Yakubovich, 2019).[2] Therefore, identification of this underlying causal schema has significant implications for improving HR practices involving selection, compensation, and other people-centric organizational practices.

**3.3. Protocol for participants:**

We follow the same protocol as Gick and Holyoak (1980;1983) to verify whether participants apply the analogy under 'no hint' and 'hint' conditions. Of course, we modify the protocol slightly to account for our inclusion of complexity (two stories). We led our participants to solve the problems in the target domain in the stepwise process:

**i) Exposure to stories:** We begin by providing the participant with each story in turn. We instruct the participant to read the stories carefully, summarize them, and rate the story's *ease of understanding* and *plausibility* using a 5-point Likert scale. We allow 3 minutes per story and randomize the order in which the stories are presented.

**ii) 'No hint' condition:** We present the participant with a problem-solving task, with their goal being to generate as many solutions as possible. The subjects are randomly allocated to either the 'Factory' problem or the 'HR' problem. We allow the participants 6 minutes to solve the problem and ask them to write down their solutions as text (we provide them the option of providing multiple solutions in multiline text box).

**iii) 'Hint' condition:** After the 6 minutes have elapsed, we explicitly offer the following hint: *"You can use one of the stories from the first text summarization exercise to solve the puzzle."* Post this prompt, we allow 3 additional minutes for the participant to re-write their solution to the problem.

---

[2] Also see: https://knowledge.insead.edu/leadership-organisations/enlightened-randomness by Puranam & Sen (2019).

**iv) GPT-4:** We follow the same chronology for each independent run of GPT-4. However, for the AI, we did not have to monitor the time limits to read and summarize the stories (in source domain) or generate a solution to the problems (in the target domain) as we did for the human participants.

### 3.4. Sample

For the purposes of the study, we conducted an experiment using human and AI subjects. Human subjects ($n = 210$) were undergraduate students taking the core strategy class in a leading business school based in France. After applying the pre-registered attention check (minimum time to complete the survey $t_{min}=3$ minutes and including only fully completed surveys) and removing two invalid responses (two subjects pasted their answers from an LLM) we excluded eleven participants. The final sample consisted of $n = 199$ participants: with 101 (50.8%) male, 93 (46.7%) female, and 5 (2.5%) subjects who didn't indicate their gender. AI sample ($n = 60$) comprised of independent runs of GPT-4 mirroring the procedure with human participants (GPT-4 capability as of September 2024). By "independent run", we man that each instance of GPT-4 was re-initiated from a new state to prevent context carryover. It is worth highlighting that, per problem, $n = 60$ would mean 15 observations per condition (hint/no hint vs. story1/story2). Due to little variation in the AI output to the prompt, 15 observations per condition were deemed sufficient.

### 3.5. Measuring "correct analogy" (causal mapping)

We pre-defined correct analogical transfer for the two stories using Directed Acyclic Graphs (DAGs) that represent the underlying causal schemas connecting the source and target domains. Figure 1 (a), 1 (b) display the DAGs for both stories alongside their corresponding target problems used in our experiment, respectively. We then manually analyzed participants' textual responses to these problems and coded a response as demonstrating "correct analogical transfer" if it reflected that the accurate abstracted schemas, as per the DAGs, were used to solve the relevant problem.

**Figure 1 (a). DAG for Radiation Story to City Factory Problem**

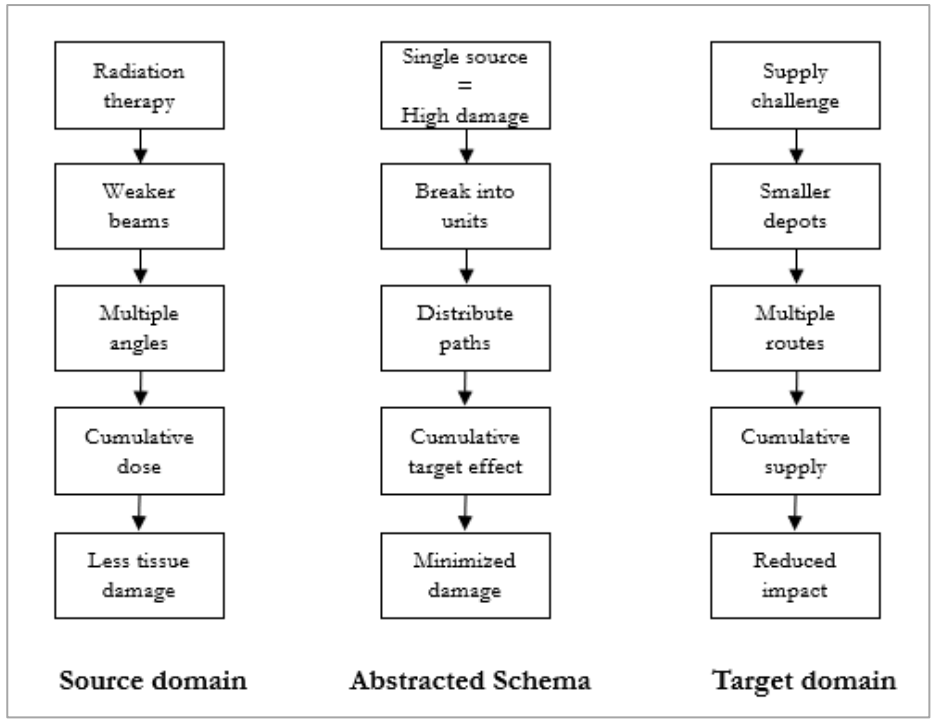

**Figure 1 (b). DAG for Dolphin Story to HR Problem**

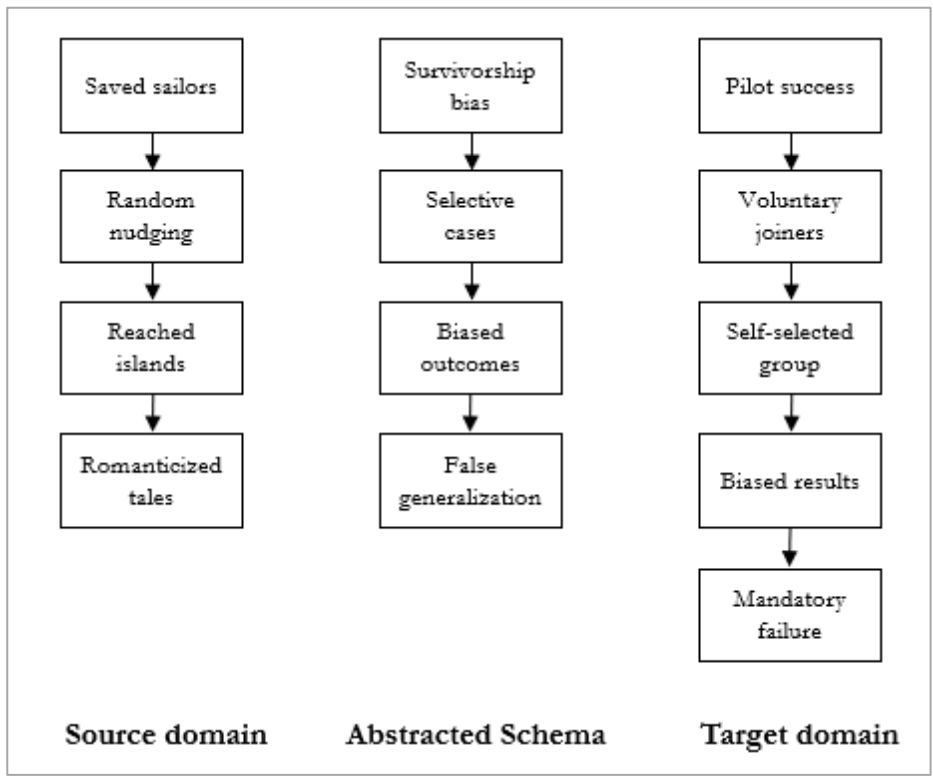

## 4. RESULTS

### 4.1. Ease of understanding

First, following standard protocol, the subjects were asked to summarize each story in a few words and evaluate the ease of understanding and plausibility (how likely they considered the story to be) (e.g., Gick & Holyoak, 1983). Ease and plausibility were evaluated on a five-point Likert scale, from 1 (least easy/plausible) to 5 (very easy/plausible).

Human subjects evaluated the dolphins story as easier to understand but less plausible than the radiation problem. However, AI (GPT-4) evaluated the radiation problem as both easier to understand and more plausible than the dolphin story. It is important to note that, for the AI case, both the ease of understanding and plausibility may reflect, in part, the semantic similarity of the stories' text within the model's training dataset. For instance, the radiation problem (established in the 1980s to study analogical reasoning), is likely to have a higher probability of featuring in the AI's training set as opposed to the dolphin story, which we have specifically created for this study. Table 1 presents a summary of these results.

**Table 1. Comparison of the evaluations of easiness and plausibility**

| Story | Human subjects | | GPT-4 | |
|---|---|---|---|---|
| | Ease | Plausibility | Ease | Plausibility |
| Radiation | 3.905 (.988) | **4.075 (.894)** | **4.500 (.504)** | **4.650 (.481)** |
| Dolphins | **4.473 (.679)** | 3.522 (.913) | 4.333 (.475) | 4.083 (.279) |
| Difference | -0.567** | 0.552** | 0.167* | 0.467** |

Note: * p(T > t) < 0.01   ** p(T > t) < 0.001; higher values in bold.

### 4.2. Human and AI problem solving accuracy

We analyzed the responses by looking at the evidence of having used an analogy. First, we found variance in the solvability of the two problems. With or without hint, human subjects used the correct analogy to solve the HR problem more readily than the Factory problem. Second, along expected lines, providing a hint to use an analogy increased its use for both problems. While 8.7% (4 out of 46) of respondents solved the factory problem after giving a hint (vs. 6.3% without a hint), nearly 37% (17 out of 46) of respondents solved the HR problem after a hint (vs. 20.3 % without a hint). Third, the solvability of both the factory and HR problems was lower than the solvability of the original problems in the classical Gick & Holyoak (1983, Experiment 2) study, where 29% (8 out of 28) of respondents solved the problem without a hint and 50% (14 out of 28) with a hint. These insights point to two things: a) the solvability of problems (with or without a hint to use an analogy) has some element of problem or context-specificity. In this case, survivorship bias was easier to spot and solve for than the operational puzzle posed in the factory problem, and b) the lower overall solvability of our study compared with the original studies alludes to the matching problem – the mere fact that there was an element of matching from two source stories to two target problems in our setup, compared to a one to one analogous transfer in the original studies, could account partly for the difference between our study and the original studies.

As for the LLM, the results show a huge effect of providing a hint for the Factory problem (Radiation story). Without a hint, only 6.7% (1 out of 15) subjects used an analogy, while with a hint, all subjects (15 out of 15) provided an analogy. In the case of the HR problem (Dolphins story), even without a hint, almost in all cases (14 out of 15) AI agent used an analogy. With a hint, all subjects (15 out of 15) used an analogy.

Figures 2 (a) and 2 (b) below show the proportion of human subjects and GPT-4 trials that solved the City factory and HR problems correctly with and without a hint, respectively.

**Figure 2 (a). Human subjects**

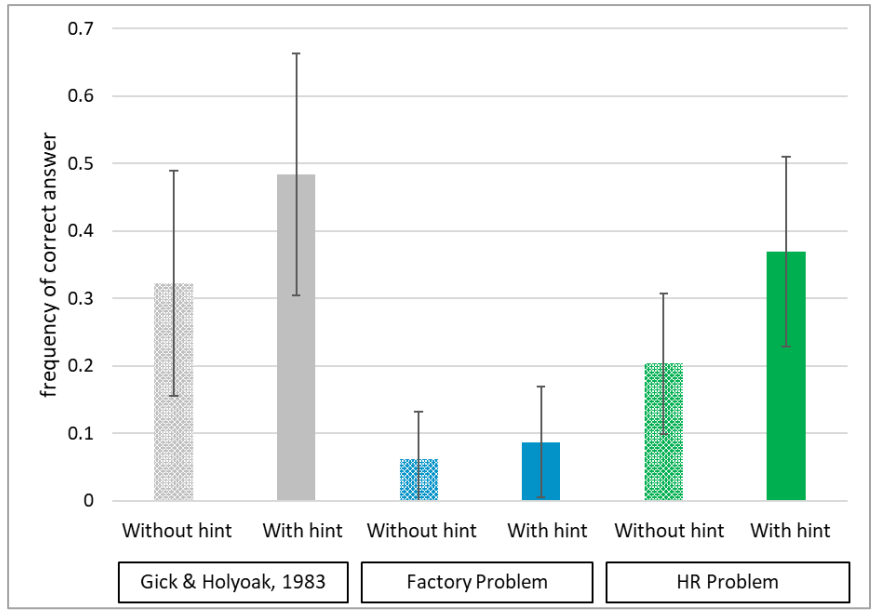

Note: Whiskers indicate 95% confidence interval

**Figure 2 (b). AI trials (GPT-4)**

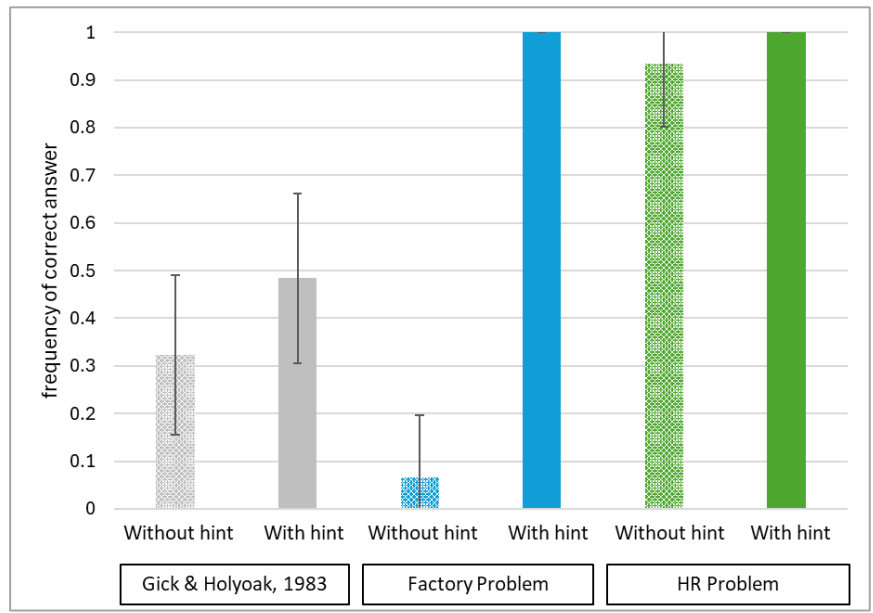

Note: Whiskers indicate 95% confidence interval

## 4.3. Application of correct analogy

Applying an analogy to solve a problem does not automatically signify success when there are many potential sources of analogies. Using the human and AI responses, we analyze the accuracy of structural mapping of both systems using a confusion matrix analysis. We look at both cases, where subjects were not provided with a hint and the cases where they were provided with a hint to use an analogy. For each case we have four confusion matrices for a combination of subject (Human & AI) and source (Dolphin & Radiation). Each confusion matrix shows whether a given story was a source of correct analogy for the given problem and whether the subject used it. The tables below show the distribution of responses over these confusion matrices.

We begin with a case without a hint. Table 2(a) through Table 2(d) show the respective results.

Table 2 (a). Claimed use of the Radiation story by an AI Agent

|  | Claimed to use radiation story | Did not claim to use radiation story |
|---|---|---|
| Truth: Radiation story is the right analogy | 0 | 15 |
| Truth: Radiation story is not the right analogy | 0 | 15 |

Table 2 (b). Claimed use of the Dolphin story by an AI Agent

|  | Claimed to use dolphin story | Did not claim to use dolphin story |
|---|---|---|
| Truth: Dolphin story is the right analogy | 0 | 15 |
| Truth: Dolphin story is not the right analogy | 0 | 15 |

**Table 2 (c). Claimed use of the Radiation story by a Human Agent**

|  | Claimed to use radiation story | Did not claim to use radiation story |
|---|---|---|
| Truth: Radiation story is the right analogy | 3 | 45 |
| Truth: Radiation story is not the right analogy | 4 | 55 |

**Table 2 (d). Claimed use of the Dolphin story by a Human Agent**

|  | Claimed to use dolphin story | Did not claim to use dolphin story |
|---|---|---|
| Truth: Dolphin story is the right analogy | 13 | 46 |
| Truth: Dolphin story is not the right analogy | 0 | 48 |

Synthesizing the above four confusion matrices, we get the following distribution of True Positives (TP), False Positives (FP), True Negatives (TN), and False Negatives (FN) (see Table 3).

**Table 3. Summary results for the without-hint condition**

| Case | TP | FN | FP | TN |
|---|---|---|---|---|
| AI: Radiation | 0 | 15 | 0 | 15 |
| AI: Dolphin | 0 | 15 | 0 | 15 |
| Human: Radiation | 3 | 45 | 4 | 55 |
| Human: Dolphin | 13 | 46 | 0 | 48 |

This distribution allows us to calculate four critical metrics that would reveal meaningful insights on the efficacy of analogical transfer across stories/problems, for human decision makers and AI:

- Precision = TP / (TP + FP)

- Recall = TP / (TP + FN)

- F1 score = 2 × (Precision × Recall) / (Precision + Recall). This statistic provides an overall performance by taking into account both precision and recall. If F1 is 1, the agent exhibits perfect precision and recall. When it is 0, either precision or recall are 0.

- Accuracy = (TP + TN) / (TP + TN + FP + FN)

Applying these formulae, we arrive at the following four measures for the four cases (agent x analogy) presented in Table 4.

**Table 4. Summary metrics for claiming the correct source story (without a hint)**

|  | Precision | Recall | F1 score | Accuracy |
|---|---|---|---|---|
| AI: Radiation | 0.00 | 0.00* | 0.00 | 0.50 |
| AI: Dolphin | 0.00 | 0.00* | 0.00 | 0.50 |
| Human: Radiation | 0.42 | 0.06 | 0.11 | 0.54 |
| Human: Dolphin | 1.00 | 0.22 | 0.36 | 0.57 |

Note: * = 0/0 → undefined, usually noted as zero

These results reveal a crucial difference in how Human and AI agents apply analogical transfer to problem solving when not given an explicit hint to use prior stories. In both cases (Radiation and Dolphin story), the AI agents failed to identify and apply the correct analogy even once, resulting in zero precision, recall, and F1 scores, though achieving 50% accuracy due to correctly rejecting incorrect analogies. In contrast, human participants showed limited but real success: in the Radiation case, they achieved perfect precision (i.e., when they claimed to have used the analogy, it was the correct one), but recall was very low (6.25%), indicating that most failed to apply the analogy at all. In the Dolphin case, humans achieved higher recall (22%) with perfect precision (100%), suggesting better—but still modest—use of the correct story. Overall, humans outperformed AI in identifying and applying correct analogies, though both groups struggled, highlighting the difficulty of spontaneous analogical reasoning across problems.

We then perform the same analysis for the condition with a hint. This reveals a stark difference in how a hint affects each group of subjects. Table 5 and Table 6 show the summary results and metrics for that case.

**Table 5. Summary results for the with-hint condition**

|                  | TP | FN | FP | TN |
|------------------|----|----|----|----|
| AI: Radiation    | 15 | 0  | 14 | 1  |
| AI: Dolphin      | 15 | 0  | 15 | 0  |
| Human: Radiation | 12 | 34 | 6  | 40 |
| Human: Dolphin   | 18 | 28 | 6  | 40 |

**Table 6. Summary metrics for claiming the correct source story with-hint**

|                  | Precision | Recall | F1 score | Accuracy |
|------------------|-----------|--------|----------|----------|
| AI: Radiation    | 0.52      | 1.00   | 0.68     | 0.53     |
| AI: Dolphin      | 0.50      | 1.00   | 0.67     | 0.50     |
| Human: Radiation | 0.67      | 0.26   | 0.38     | 0.57     |
| Human: Dolphin   | 0.75      | 0.39   | 0.51     | 0.63     |

These results under the with-hint condition show that humans remain conservative in their application of analogy while for AI, a hint significantly improves its ability to identify and apply the correct analogy. In comparison, human performance improves more modestly. For both AI cases (Radiation and Dolphin story), recall remains perfect (1.00), meaning the AI subjects always used the correct analogy when it applied, but precision hovers around 0.50 due to frequent false positives. This leads to moderate F1 scores (0.67–0.68) and slight improvements in accuracy (50–53%). Human subjects, by contrast, show better precision (0.67–0.75), indicating more selectivity and fewer false claims, but their recall is lower (0.26–0.39), suggesting they still often miss the correct analogy even when prompted. Human subjects' F1 scores (0.38–0.51) and accuracy (57–63%) are improved compared to the no-hint condition but remain behind AI in recall-driven metrics. This pattern suggests the AI is over-flagging to avoid missing critical positives (evidence of the "demand effect" where LLMs modify their behavior based on what they believe researchers want to see (Gui & Toubia, 2023). Table 7 further synthesizes these differences along the four metrics .

**Table 7. Comparison and summary of the result**

| Metric | AI (No Hint) | Humans (No Hint) | AI (Hint) | Humans (Hint) | Pattern | Dominant Error Type |
|---|---|---|---|---|---|---|
| Precision | None (0.0) | Moderate to Perfect (0.42–1.0) | Moderate (0.50–0.52) | High (0.67–0.75) | AI improves sharply with hint | AI: Inactive → Over-diagnosis |
| Recall | None (0.0) | Poor (0.06–0.22) | Perfect (1.0) | Low to Moderate (0.26–0.39) | Humans benefit less from hint | Humans: Consistent under-diagnosis |
| F1 Score | None (0.0) | Low (0.11–0.36) | Moderate (0.67–0.68) | Modest (0.38–0.51) | F1 increases in all cases | F1 gains driven by recall in both |
| Accuracy | Poor (0.5) | Modest (0.54–0.57) | Slightly better (0.50–0.53) | Slightly better (0.57–0.63) | Small accuracy gains with hint | Mostly balanced across conditions |
| Error Type | Inactivity (no response) | Under-diagnosis (false negatives) | Over-diagnosis (false positives) | Continued under-diagnosis | AI flips strategy with hint (demand effect) | AI: FP increase; Humans: FN remain |

The summary reveals a clear difference in how AI and human subjects respond to hints when applying analogies to strategic problems. Without a hint, AI fails to recognize or apply analogies, resulting in zero precision, recall, and F1 score—essentially complete absence of the evidence for analogical reasoning. Human subjects, on the other hand, demonstrate high precision but very low recall, indicating that they only claim to use analogies when highly confident, often missing to spot an opportunity for analogical transfer (under-diagnosis). When given a hint, AI performance improves dramatically in recall (up to 1.0) but suffers from over-application of analogies, leading to moderate precision and many false positives (over-diagnosis). Human subjects also improve with hints, particularly in recall, but still lag behind AI in overall analogy detection. Their under-diagnosis pattern persists, which suggests that hints are not sufficient to overcome their conservative approach. Thus, our results indicate that providing a hint helps both groups, but AI shows a more dramatic reaction, which introduces a new trade-off between correct and incorrect analogy application.

The findings suggest that AI and human decision makers could be complementary in analogical reasoning to solve strategic problems. Current AI systems, when given a hint, demonstrate high recall—they rarely miss opportunities to apply analogies—but are prone to overapplication; often suggesting analogies that do not apply. In contrast, human decision-makers show high precision, typically identifying analogies correctly, but suffer from conservative approach; frequently failing to recognize relevant analogies–even when prompted. These patterns suggest a potentially productive division of labor in organizational decision-making that requires analogical transfer: AI can serve as an analogy generator, surfacing potential analogical matches across diverse contexts, while human decision makers act as critical evaluators, exercising contextual judgment to assess which analogies are truly applicable.

### 4.4. Nature of misappropriation of analogy (Surface vs. Structural similarity)

Figures 3 (a), 3 (b) illustrate the nature of incorrect analogical transfer—i.e., misappropriation of analogies—by GPT-4 and human subjects, respectively. On the unconditional set of responses (i.e., hint or without hint), we first code each response as an instance of surface and structural similarity between the source stories and target problems based on the definition of "surface" and "structural" match of Holyoak and Koh (1987), and Gentner and Smith (2013).

We define incorrect analogical transfer as instances where participants applied the radiation story to solve the HR problem or the dolphin story to solve the city factory problem. Within this misapplication, we categorize the responses based on whether the mistaken analogy transfer was *initiated* due to surface similarity (e.g., dolphin→sea (the common feature)→city factory with port) or a "wrong" causal structure with respect to our pre-defined "correct" DAGs. In some cases, AI drew analogies from both stories to solve a particular problem, using a mix of incorrect surface and structural analogies from either story (cases coded as "both"). Effectively, these cases may represent partially correct analogical transfer. However, the propensity for AI to use both stories also highlights the 'demand effect'—i.e., the LLM's propensity to find analogies from both stories just because those stories are available and must be utilized regardless of the degree or nature of the similarity (structural or surface/superficial).

**Figure 3 (a). Nature of false analogy by AI**

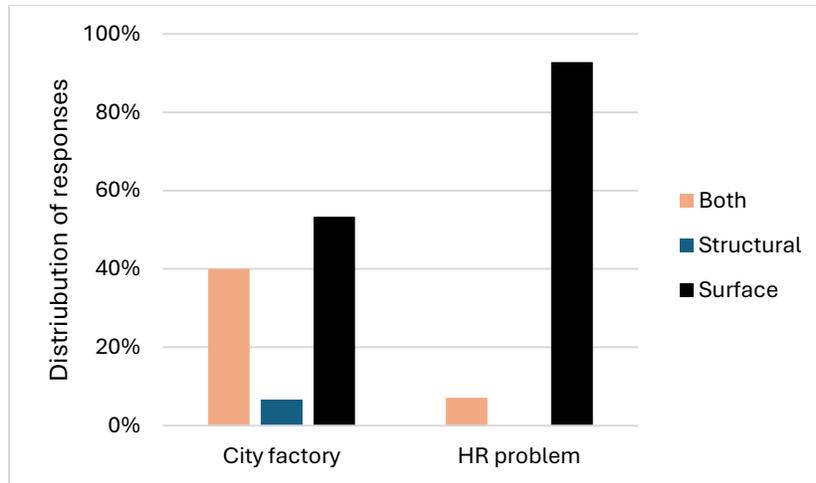

**Figure 3 (b). Nature of false analogy by Humans**

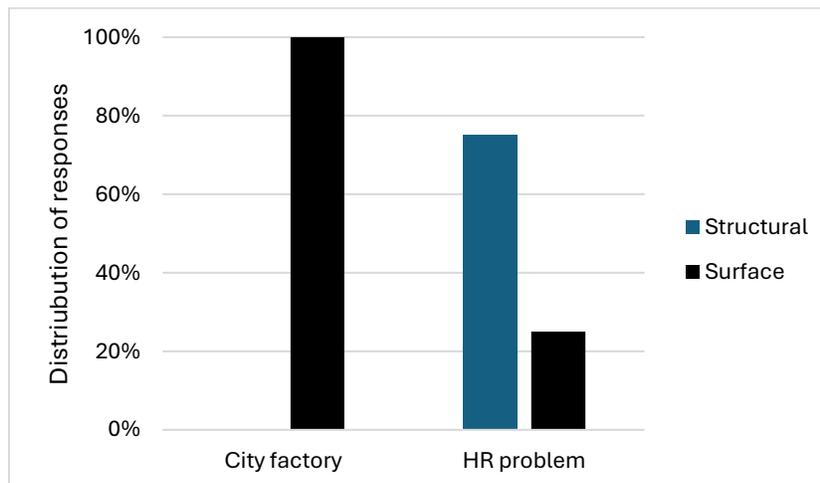

These findings are revealing of systematic patterns in erroneous analogical transfer. GPT-4's false analogies are overwhelmingly driven by surface-level similarities—semantic overlap, contextual cues, or textual proximity—rather than deep structural mappings. In contrast, when human participants misapply an analogy, they are more likely to incorrectly transfer an abstract causal schema, suggesting a misfiring of structural mapping rather than superficial matching. This contrast underscores a fundamental difference in analogical reasoning errors: AI overgeneralizes based on surface patterns, while humans overreach on inferred causal logic.

Table 8 gives an example each from AI and Human doing surface versus structural misappropriations of analogical transfer.

**Table 8. Examples of surface versus structural misappropriations of analogical transfer**

| Subject | Problem | Type of mistake | Excerpts from responses |
|---|---|---|---|
| GPT-4 | HR | Surface | "Dr. Clarke's story about radiation therapy is a good analogy for the need for precision. Just as a one-size-fits-all approach in radiation therapy can harm healthy tissues, a blanket training program for everyone might not be optimal. The voluntary nature of the pilot meant that those who felt it wouldn't benefit them could opt out." |
| GPT-4 | City factory | Surface | "Drawing inspiration from the dolphin story, it's worth noting that nature sometimes offers unexpected solutions. In this context, consider researching and investing in underwater drone technology for goods transportation." |
| GPT-4 | City factory | Structural | "Local Distribution Centers (Inspired by the sailor's survival stories) Build a local distribution center outside the city, in a more accessible location. Goods from the factory can be transported to this center in small batches during non-peak hours." |
| Human | HR | Surface | "Offer not so many places to the experiment, let's say only 10 places by month (for a 200 employees company). And in the same time, communicate about result for the participant, in a scientific way." |
| Human | HR | Structural | "Just like the case with Dr. Clarke, where one strong radiation beam cured cancerous tissues but in turn had collateral damage on surrounding tissues, although the training program worked for a section of the organization, it also negatively affected others to the point the overall impact was for all practical reasons, negative (declined productivity)." |
| Human | City factory | Surface | "Explore sustainable transport : by inspiring yourself from animals in the sea" |

These results suggest that AI tools like GPT-4 may appear confident and fluent in drawing analogies, but their reasoning often lacks depth as of today. Managers relying on AI outputs for decision support must remain cautious, especially when analogies seem intuitively appealing. In contrast, human reasoning, while more cautious, is vulnerable to overextending causal theories. In high-stakes strategic settings, the optimal approach may lie in combining AI's breadth of suggestion with human judgment's ability to vet causal validity—thereby mitigating both over-diagnosis and under-diagnosis in analogical reasoning.

## 5. DISCUSSION & CONCLUSION

This study set out to evaluate whether a state-of-the-art large language model (GPT-4) can match or surpass human reasoners in the *matching* phase of analogical reasoning when tasks are framed entirely in natural language and embedded within managerial contexts. To this end, we employed a $2 \times 2$ experimental design in which participants—either business school students (n = 199) or GPT-4 instances (n = 60)—were presented with two candidate source analogies (radiation therapy; survivorship bias) and tasked with identifying the more structurally appropriate analogue for solving one of two novel target problems (factory logistics; HR pilot study). All conditions were held constant across human and machine participants, save for the agent performing the task.

Four robust patterns emerged from the data. First, the introduction of a source *selection* step—requiring participants to choose between competing analogues—substantially lowered task solvability relative to traditional one-to-one analogical transfer paradigms. This confirms that source matching imposes a nontrivial cognitive burden, consistent with prior theoretical treatments that distinguish retrieval and evaluation from structural mapping.

Second, in conditions where the existence of an appropriate source was explicitly pre-selected (the "hint" condition), GPT-4 consistently outperformed human participants in identifying and mapping the relevant structural correspondences. This suggests that the stimulus to retrieval acts more effectively on GPT-4 than it does on our human subjects.

Third, human and machine reasoners exhibited sharply divergent error profiles when asked to explain the mapping they employed between source and target. Human participants achieved high *precision* (up to 0.75), meaning that when they selected a source, it was always correct—but at the cost of low *recall* (as low as 0.26), indicating a high rate of non-selection even when a correct match was available. In contrast, GPT-4 displayed the reverse pattern: it achieved perfect recall, successfully identifying all cases of structurally valid sources, but at the expense of lower precision ($\approx 0.50$), generating a high number of false positives in the process.

Finally, the nature of these errors differed systematically across entities. GPT-4's false positives were predominantly driven by *overgeneralization from surface-level features*, consistent with its training regime and known limitations in discerning deep relational structure. By contrast, human errors typically stemmed from *misinterpretations of causal structure*—suggesting a higher threshold for source endorsement but also a greater susceptibility to misconstrue when analogical depth is obscured.

**Implications for the Theory of Analogical Reasoning in Strategic Decision Making**

While analogical reasoning has long been recognized as central to strategic cognition, most extant models emphasize retrieval and mapping, drawing heavily on insights from cognitive psychology. The third step—*evaluation*, or the filtering of candidate analogues for structural coherence—has remained analytically underdeveloped despite its salience in practice. Our findings suggest that this omission is consequential. In particular, the matching stage—where one must adjudicate among superficially plausible analogues—emerges as the locus at which human strategic reasoners retain a decisive advantage over state-of-the-art large language models. The contrast in error profiles is telling: human participants exhibited high precision and low recall, a pattern aligned with a "hedgehog" cognitive style (Tetlock, 2017) that privileges cautious commitment and minimizes false positives (Type I errors). GPT-4, by contrast, adopted a more "fox-like" stance—exhibiting high recall and lower precision, thereby reducing false negatives (Type II errors) at the cost of admitting spurious analogies. These divergent tendencies underscore the value of elevating *matching* to a first-order construct in theories of strategic analogy, thus extending the canonical dyad of retrieval and mapping into a triadic model that better reflects how analogical reasoning unfolds in open-ended problem settings.

More broadly, these results invite a reconsideration of how analogical reasoning functions as a search heuristic in ill-structured strategic environments. Large language models demonstrably expand the *breadth* of cognitive search by retrieving a far wider range of analogical sources than human reasoners typically access unaided. Yet this expansion comes with an evaluative cost: the inability to systematically vet analogues for causal isomorphism. In contrast, human reasoners—while far more limited in analogical recall—exhibit a marked advantage in *depth*, particularly in their capacity to impose relational and causal structure on candidate mappings.

The distinct error profiles of humans and LLMs further provide micro-foundational insight into the potential for human–AI complementarity in strategic decision-making contexts. A possible division of labor with specialization may involve LLMs cast as high-throughput *generators* of plausible analogical frames, expanding the strategic option set through computational breadth. Human agents, in turn, function as *evaluative filters*, imposing coherence by discriminating among analogues based on deeper structural alignment. However, if human skills at retrieval are to be preserved, one might instead adopt an ensemble approach (Choudhary et al., 2025), where humans and LLMs independently retrieve and map analogies, with a comparison and selection between the final outputs of the two. By experimentally demonstrating that LLMs and humans err in systematically different ways, our findings lay the groundwork for more fine-grained design of decision workflows in strategy that explicitly leverage this asymmetry—allocating analogical generation to machines, while preserving final analogical selection as a human prerogative.

**Implications for Managerial Practice**

The findings carry several actionable implications for the design and governance of AI-assisted strategic decision-making. First, organizations deploying large language models should treat model outputs not as definitive recommendations but as *candidate sets*—preliminary suggestions whose utility depends on subsequent human filtering. Interfaces that render the underlying causal schemas more transparent—such as through visualizations of structural alignments (e.g., directed acyclic graphs)—can support this division of labour by combining algorithmic breadth with human depth in evaluation.

Second, the persistence of false negatives among human participants—driven by misidentification of the abstract causal structure—underscores the continued importance of training managers in structural vigilance. The advent of LLMs does not reduce the need for schema-based reasoning; rather, it amplifies its significance. As machines expand the analogical search space, the human task becomes increasingly one of discerning which structural correspondences truly warrant attention. Thus, training that emphasizes systematic comparison, causal mapping, and analogical discrimination remains essential.

Third, our results suggest that managerial reliance on either humans or machines should be guided by an explicit awareness of *error cost asymmetries*. In decision contexts where the cost of missing a valid analogy outweighs the cost of entertaining false leads—such as in time-sensitive crisis response—leveraging LLMs' high recall may be desirable. Conversely, in high-stakes strategic decisions where false positives carry irreversible consequences—such as major capital investments or organizational pivots—human filtering, with its higher specificity, should dominate.

Finally, the findings support a governance architecture centered on *human-in-the-loop* oversight. Delegating analogical reasoning wholesale to LLMs—particularly when their outputs are driven by superficial feature similarity—risks systematic misframing of strategic problems. Embedding human judgment at critical evaluation junctures serves not only as a safeguard against such misalignment but also as a means of realizing the potential complementarities between machine-generated option sets and human causal reasoning.

**Limitations and Avenues for Future Research**

As with any experimental study, certain limitations constrain the scope of our findings and point toward promising directions for future inquiry. Our human participant pool consisted of business school students—a common proxy for boundedly rational decision-makers in strategic cognition research, but nonetheless a sample of strategy novices. While appropriate for isolating cognitive mechanisms, replication with senior executives or cross-cultural cohorts would enhance the external validity of our claims. The task environment itself was deliberately simplified: participants were asked to select between two analogical sources for each of two target problems. While this structure enabled tight control over the matching challenge, it does not fully reflect the complexity of real-world strategic decision-making, where the analogical search space may contain dozens of plausible comparisons. Scaling the number of candidate sources and systematically varying the search costs could illuminate how human and AI capabilities interact under conditions of greater ecological realism.

Additionally, our focus on a single model—GPT-4 (as of September 2024)—captures a frontier instantiation of transformer-based architectures but does not exhaust the design space. Comparative evaluations across model families, sizes, and training objectives are needed to assess whether the high-recall/low-precision profile observed here generalizes beyond this specific implementation. Moreover, our use of a static, single-shot prompt limits the capacity for dynamic engagement. Iterative prompting—where the model can query constraints or receive feedback—may enable more precise identification of structural correspondences and deserves empirical exploration.

Finally, while we employed directed acyclic graph (DAG) representations and intercoder reliability checks to evaluate the quality of analogical transfers, the assessment of structural fit still involved a degree of qualitative judgment. Incorporating automated schema-matching techniques or embedding model-generated rationales into the evaluation process could complement human coding and enhance reproducibility in future datasets. Collectively, these extensions would not only assess the robustness of the current findings but also help refine theoretical and practical understandings of human–AI complementarity in analogical reasoning.

**Conclusion**

Notwithstanding the limitations outlined above, this study offers three contributions to the literature on strategic cognition. First, it foregrounds *matching*—the evaluative act of selecting structurally appropriate analogies—as a central and distinct component of analogical reasoning in strategic decision-making. While prior work has emphasized retrieval and mapping, our findings suggest that it is in the evaluative phase where reasoning quality is most meaningfully differentiated, and where human expertise retains an enduring comparative advantage. Second, the study provides the first direct empirical evidence that large language models and human strategists exhibit *complementary but asymmetric* analogical capabilities: LLMs expand the analogical search space through high recall, while humans contribute discriminative value through high precision. This asymmetry is not a liability but a design opportunity.

Third, we translate this insight into practical implications for the architecture of AI-assisted decision-making processes. Rather than replacing human strategic reasoning, LLMs may serve as high-throughput generators of candidate analogies, with human agents functioning as evaluative filters to preserve causal coherence. As language models continue to evolve in scale and fluency, the central challenge will not be whether they can reason analogically, but under what conditions their broad but noisy suggestions can be productively integrated with the narrower but more causally grounded judgments of human decision-makers. Designing decision workflows that systematically leverage this complementarity constitutes a critical frontier for both future research and managerial practice.